\documentclass{article}

\usepackage{arxiv}

\usepackage[utf8]{inputenc} 
\usepackage[T1]{fontenc}    
\usepackage{hyperref}       
\usepackage{url}            
\usepackage{booktabs}       
\usepackage{amsmath}
\usepackage{amsfonts} 
\usepackage{nicefrac}       
\usepackage{mathabx}
\usepackage{microtype}      
\usepackage{lipsum}
\usepackage{graphicx}
\usepackage{float}
\usepackage{todonotes}

\newcommand{\specialcell}[2][c]{%
  \begin{tabular}[#1]{@{}c@{}}#2\end{tabular}}

\DeclareMathOperator*{\argmin}{arg\,min}

\usepackage{amsthm}
\theoremstyle{definition}
\newtheorem{definition}{Definition}[section]

\RequirePackage{algorithm}
\RequirePackage{algorithmic}
\title{Lifting Interpretability-Performance Trade-off via Automated Feature Engineering}

\author{
  Alicja Gosiewska \\
  Faculty of Mathematics and Information Science\\
  Warsaw University of Technology\\
  \texttt{a.gosiewska@mini.pw.edu.pl} \\
   \And
   \And
 Przemyslaw Biecek \\
    Faculty of Mathematics, Informatics and Mechanics\\ 
    University of Warsaw \\
  Faculty of Mathematics and Information Science\\
  Warsaw University of Technology\\
  \texttt{przemyslaw.biecek@gmail.com} \\
}

\begin{document}
\maketitle

\begin{abstract}
Complex black-box predictive models may have high performance, but lack of interpretability causes problems like lack of trust, lack of stability, sensitivity to concept drift. On the other hand, achieving satisfactory accuracy of interpretable models require more time-consuming work related to feature engineering. Can we train interpretable and accurate models, without timeless feature engineering? 
We propose a~method that uses elastic black-boxes as surrogate models to create a simpler, less opaque, yet still accurate and interpretable glass-box models. New models are created on newly engineered features extracted with the help of a~surrogate model.
We supply the analysis by a~large-scale benchmark on several tabular data sets from the OpenML database. There are two results 1)~extracting information from complex models may improve the performance of linear models,  2)~questioning a common myth that complex machine learning models outperform linear models.  
\end{abstract}

\section{Introduction}
\label{motivation}

Data preparation and transformations are at the core of most data analyses. The quality of the algorithm is dependent not only on its complexity but also on the features engineering step. Understanding the data and confronting it to the domain knowledge is crucial in finances, insurance, medical field and many others. 
Although several automated data transformation techniques have been
developed and improved in performance, they remain time-consuming and often worse than manual human work. As a~result, years of theoretical and empirical work has attempted to generate automated data transformations. One of the most common are Principal Component Analysis or Factor Analysis. More sophisticated approaches to automated feature engineering are based on copulas \citep{7344858}, iteratively generated non-linear features \citep{horn2019autofeat}, or even training neural networks to predicting the transformations impact \citep{inproceedings}. Yet still newly produced features are difficult to interpret. The lack of interpretablility leads to the lack of trust in model's predictions.

\begin{figure}[t]
    \centering
    \includegraphics[width=0.6\textwidth]{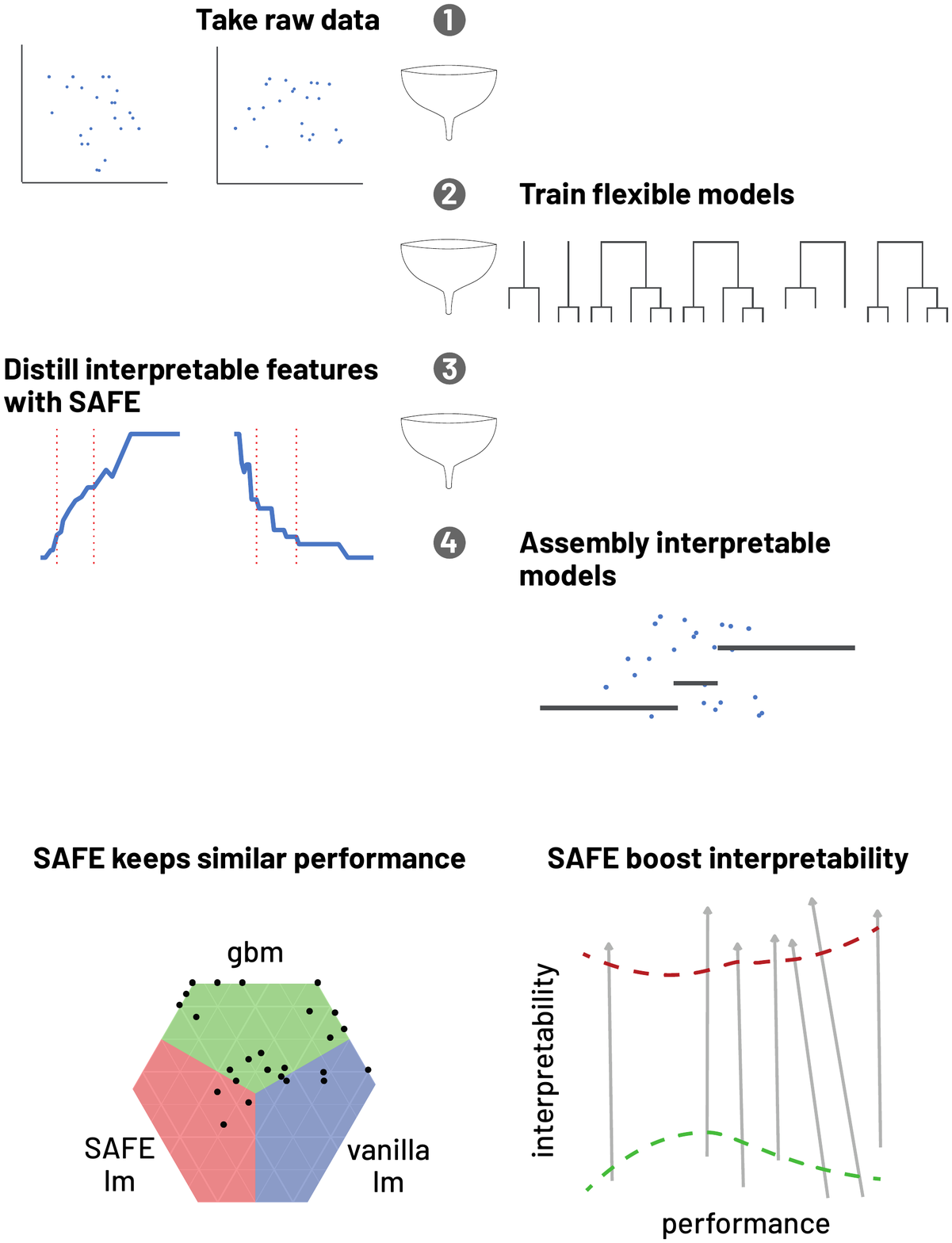}
    \caption{The process of the navigated distillation of information for raw data with the use of flexible models into an interpretable model. Each step is connected with decreasing of the complexity. }
    \label{fig:Diagram2}
\end{figure}

Questions of trust in machine learning models became crucial issues in recent years. Complex predictive models have various applications in different areas \cite{PALIWAL20092, KOUROU20158}. Hence, it is important to ensure that predictions of these models are reliable. There are four requirements whose fulfillment is essential to ensure that predictive model is trustworthy and accessible: (1) high model performance, (2) auditability, (3) interpretability, and (4)~automaticity.

Requirement 1. High model performance means that a~model rarely makes wrong predictions or the prediction error is small on average. Usually, this can be achieved by using complex, so-called black-box models, such as, boosting trees \cite{DBLP:journals/corr/ChenG16} or deep neutral networks \cite{Goodfellow-et-al-2016}. The opposite of black-boxes are glass-boxes. They are simple, interpretable models, such as linear regression, logistic regression, decision trees, regression trees, and decision rules.

Model performance ensures only a part of information about model's quality. Model requirement 2, i.e. auditability, guarantees that the model can be verified with respect to different criteria. They are, for example, stability, fairness, and sensitivity to a~concept drift. There are tools that allow to audit black-box models \cite{gosiewska2018auditor}, yet simple glass-boxes offer more extended range of diagnostic methods \mbox{\cite{Harrell:2006:RMS:1196963}.}

The third requirement is the interpretability, which became an important topic in recent years \cite{ONeil}. Machine learning models influence people's lives, in particular, they are used by financial, medical, and security institutions. Models have an impact on whether we get a~loan \cite{HUANG2007847}, what type of treatment we receive \cite{doi:10.1177/117693510600200030}, or even whether we are searched by the police \cite{4053200}. Therefore, models reasoning should be transparent and accessible. There is an ongoing debate about the right to explanation, what does it mean and how it can be achieved \mbox{\cite{DBLP:journals/corr/abs-1711-00399, Edwards_Veale_2018}.}

The (4) automaticity of machine learning methods is spreading rapidly.
Due to the increasing computational power, it becomes easier and easier to obtain more precise models, usually in an automatic manner. There are automated frameworks for AutoML like autokeras, auto-sklearn, TPOT \cite{jin2018efficient, NIPS2015_5872, Olson2016} that allow one to train a model even without any statistical knowledge or even programming skills. Yet, machine learning specialists can also take an advantage of automated methods of modeling. Such methods reduce time needed to train the model, therefore human effort can be directed towards more creative and sophisticated tasks than testing wide range of parameters and models.

People usually choose automatically fitted black-box models that achieve high performance at the cost of auditability and interpretability. 
As a response to this problem, the methodology for explaining predictions of black-box models, so called post-hoc interpretability, is under active development. There are several approaches to explaining the global behavior of black-boxes. Model can be reduced to simple if-then rules \cite{MAGIX} or decision trees \cite{proc-jsm-2018}. 
However, these explanations are simplifications of models and may be inaccurate. As a consequence, they may be misleading or even harmful. Hence, in many applications it is better to train a transparent, interpretable model than apply explanations to a complex model \cite{2017arXiv171006169T, pleseStop}. Therefore, automated methods of obtaining interpretable models, while maintaining the predictive capabilities of a~complex model, are extremely important.

In this article, we present a method for Surrogate Assisted Feature Extraction for Model Learning (SAFE). This method uses a surrogate model to assist feature engineering and leads to training accurate and transparent glass-box model. In this approach, surrogate model should be accurate to produce best feature transformation, yet it does not have to be interpretable. Based on the new features, the transparent glass-box model is trained. In many cases the high accuracy of black-box models comes from good data representation and this is something than can be next extracted from the model.
The SAFE method is flexible and model agnostic, any class of models may be used as a surrogate model and as a~glass-box model. Therefore, surrogate model may be selected to fit the data as best as possible, while glass-box model one can be selected according to the particular task or abilities of the end-users to interpreting models. 
An advantage of this methodology is that the final glass-box model has a performance close to the surrogate model. By changing the representation of the data, SAFE allows to gain interpretability with minimal or no reduction of model~performance. 
The SAFE method can be used as a~step in training a model with AutoML methods. We can use AutoML to fit elastic and complex model, then use SAFE to obtain a~transparent model.

The paper is organized as follows. Section \ref{sec:notation} contains notation and formal problem formulation.
Section~\ref{sec:SAFE_algorithm} provides a~description of the SAFE algorithm. Section~\ref{SAFE_application} contains benchmarks for the SAFE method and an example that illustrate the interpretability gain. Conclusions are in Section~\ref{discussion}.

\section{Interpretable Feature Transformations}
\label{sec:notation}

\begin{figure*}[t!h]
    \centering
    \includegraphics[width=\textwidth]{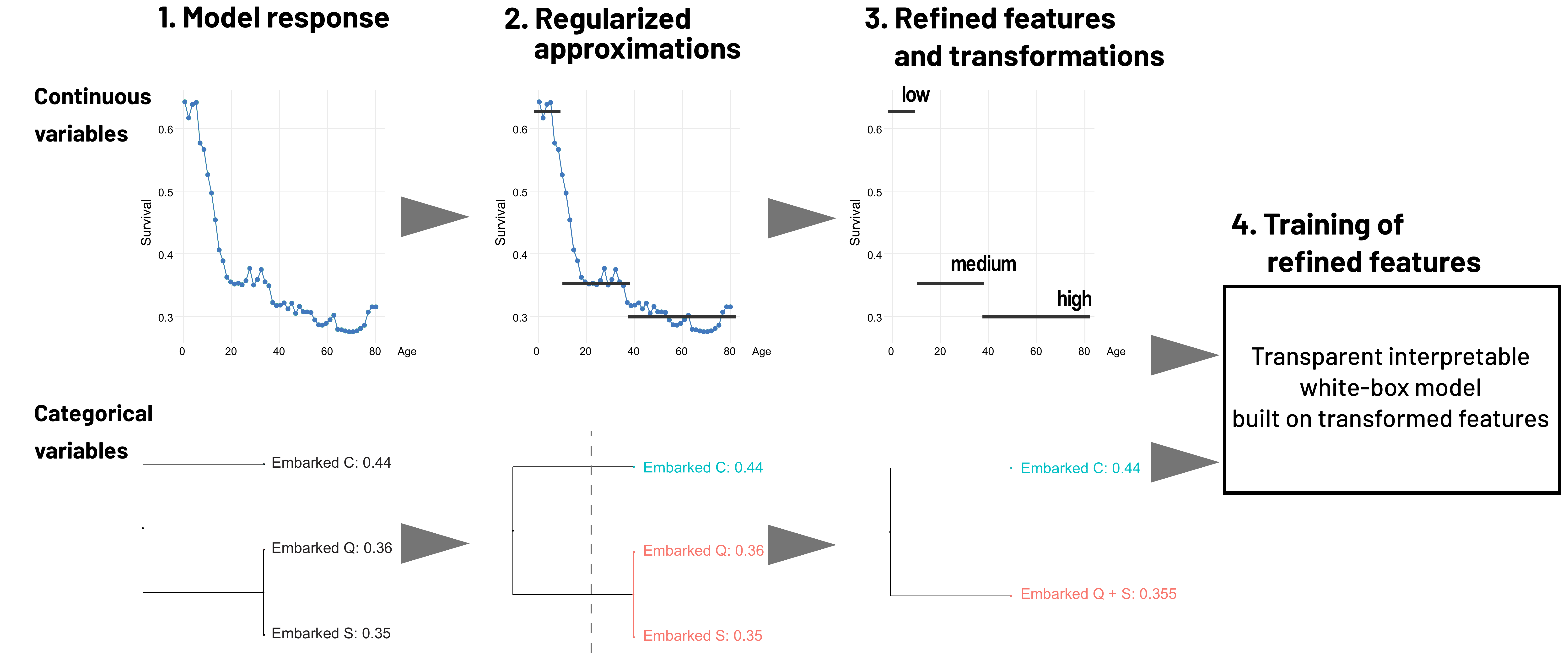}
    \caption{The SAFE algorithm in four steps, 1. train elastic surrogate model, 2. approximate model response, 3.~extract transformations and new features, 4. train refined model.}
    \label{fig:safeDiagram}
    \vskip -0.1in
\end{figure*}

The overall goal of the SAFE method is to transform original variables into new interpretable features. Now, we present formal formulation of feature transformation problem.
Let us consider a true data generating process as $m(x)$ which is a true, underlying phenomenon that is creating the data $(X, Y)$. Where $X$ is a matrix of $n$ rows (observations) and $p$~columns (independent variables) and $Y$ is a~potentially stochastic vector of $n$ response values.
We consider $X$ as a~subspace $X \subseteq \mathbb{R}^p$. Sometimes we will refer to $X$ as a~subset of the cartesian product $X \subseteq X_1 \times X_2 \times ... \times X_p$, where $X_i \subseteq \mathbb{R}$, for $i = 1,2,...,p$. Now, let $f: X \to \mathbb{R}$ be a black-box model.

Space X is dvided into aspects that will be coded separately. Coding one variable but also can be multiple variables, so we define more general way. X' is one of coded aspects.
Let $X'_K$ be a subset of $\mathbb{R}^{q_{K}}$ for some $q_K$.
We can consider vectors $x \in X$ and ${x'}_K \in {X'}_K$.

As a function $f$ represents a potentially complex model, our goal is to obtain a simple model train on the basis of knowledge gained from $f$. To accomplish this, we use relationships between variables and model response to create transformations of variables. Transformed variables may be used to train new, simple model.

Now we can define transformer functions. 
Let $h_j(x) = x'_{K_j}$, where $h_j: X \to X'_{K_j}$ be a transformer function from space $X$ into space $X_{K_j}$. 

Let $X'$ be a cartesian product of sets $X'_{K_j}$: $X' =  X'_{K_1} \times X'_{K_2} \times ... \times X'_{K_J}$. 
Now, we can define feature transformation function $h: X \to X' = X'_{K_1} \times X'_{K_2} \times ... \times X'_{K_J}$

$$h(x) = (h_1(x),h_2(x),...,h_J(x))$$

Let us note that $h_i$ could be defined on a subset of $X$ since $h_i$ do not have to include all $p$ variables. However, we set the domain of functions $h_i$ on $X$ to keep the notation as simple as possible.
Function $h$ transforms vectors from space $X$ into vectors in space $X'$.

We define a glass-box model $g: x' \in X' \to y \in \mathbb{R}$ and $g \in G$ where $G$ is a class of interpretable models.
$H$ is a~defined class of transformations.
The best glass-box model is obtained by the following formulation:
$$ g = \argmin_{g \in G} \min_{h \in H} \mathcal{L}(g(h(x)), y).$$

Where $\mathcal{L}$ is some loss function, for example, accuracy, cross-entropy, or root mean square error.
Transformations $h_i$ can be any tools to feature engineering.

We propose a novel approach SAFE in which we use partial dependence profiles and hierarchical clustering to obtain binary features that are easily interpretable. Especially when used for fitting linear models they provide easy to understand additive interpretation of model's predictions.

\subsection{SAFE as a Data-Driven Feature Transformation}
\label{sec:safe_notation}

Let us now consider transformation functions $h^{SAFE}_i: X \to \{0,1\}^{q_i}$ such as $h^{SAFE}_i$ transforms values of the $i$-th variable into binary vectors of length $q_i$.

If $x_i$ is a \textbf{categorical variable}, function $h^{SAFE}_i$ merges some of levels of $x_i$ by hierarchical clustering and find new concatenated levels. 
If $x_i$ is a \textbf{numerical variable}, function $h^{SAFE}_i$ binns $x_i$ due to the changepoints of partial dependence profile. 
Now let us introduce the partial dependence profile \cite{PDP}.

\begin{definition}{\textbf{Partial Dependence Profile (PDP)}}

Let $x_1, x_2, ..., x_p$ be features in the surrogate model $f$. A~subset of all features except $x_i$ we denote as $x_{-i}$.
The partial dependence profile is defined as
$$
f_i(x_i) = \mathbb{E}_{x_{-i}}[f(x_i, x_{-i}) ].
$$

\end{definition}

PDP is estimated as
$$
\hat f_i(x_i) = \frac{1}{n} \sum_{j=1}^{n} f(x_{i}, x_{-i}^j),
$$
where $n$ is the number of observations and  $x_i^j$ is a value of the $i$-th feature for the $j$-th instance. Partial dependence profile describes the expected output condition on a selected variable. The visualization of this function is Partial Dependence Plot \cite{RJ-2017-016}, an example plot is presented in Step~1 in Figure~\ref{fig:safeDiagram}.

A result of $h^{SAFE}_i$ transformation is a new space of interpretable binary representations of variables from a~space~$X$.



\section{Description of the SAFE Algorithm}
\label{sec:SAFE_algorithm}

The SAFE algorithm uses a complex model as a surrogate. New binary features are created on the basis of surrogate predictions. These new features are used to train a simple refined model.
Illustration of the SAFE method is presented in Figure~\ref{fig:safeDiagram}.
In the Algorithm~\ref{alg:SAFEdescription} we describe how data transformations are extracted from the surrogate model while in Algorithm~\ref{alg:SAFElearning} we show how to train a new refined model based on transformed features. The terminology being used in algorithms was introduced in Section~\ref{sec:notation} .

\textbf{The change point method} \cite{DBLP:journals/corr/abs-1801-00718} is used to identify times when the probability distribution of a time series changes.
\textbf{The hierarchical clustering} \cite{Rokach2005} is an algorithm that groups observations into clusters. It involves creating a hierarchy of clusters that have a predetermined ordering. 
Step~2 in \mbox{Figure~\ref{fig:safeDiagram}} corresponds to both change point method and hierarchical~clustering.

\begin{algorithm}[tb]
   \caption{Surrogate Assisted Feature Extraction}
   \label{alg:SAFEdescription}
\begin{algorithmic}
   \STATE {\bfseries Input:} data $X_{n \times p}$, surrogate model $M$, regularization penalty $\lambda$.
    \STATE {\bfseries Start:} 
   \FOR{$i=1$ {\bfseries to} $p$} 
   \STATE Let $x_i$ be $i$-th feature.
       \IF{$x_i$ is numerical}
       \STATE Calculate partial dependence profile $f_i(x)$ for feature $x_i$.
       \STATE Approximate $f_i(x)$ with interpretable features $x^*_{i}$, for example, use the change point method to discretize the variable with regularization penalty~$\lambda_i$.
       \STATE Save transformation $t_i(x)$ that transforms $x_i$ into~$x_i^*$.
       \ENDIF
       \IF{$x_i$ is categorical}
       \STATE Calculate model responses for each observation with imputed each possible value of $x_i$.
       \STATE Merge levels of $f_i(x)$ with similar model responses, for example use the hierarchical clustering with number of clusters $\lambda_i$.
       \STATE Save transformation $t_i(x)$ that transforms $x_i$ into~$x_i^*$.
   \ENDIF
  \ENDFOR
 \STATE Sets of transformations $T^* = \{t_1, ..., t_p\}$ may be used to create new data $X^*$ from features $x_i^* = t_i(x_i)$.
\end{algorithmic}
\end{algorithm}

\begin{algorithm}[th]
   \caption{Model Learning with Surrogate Assisted Feature Extraction}
   \label{alg:SAFElearning}
\begin{algorithmic}
   \STATE {\bfseries Input:} data $X^{new}_{m \times p}$, set of transformations $T^*$ derived from surrogate model $M$.
     \STATE {\bfseries Start:} 
\STATE Transform dataset $X$ into $X^{*, new} = T^*(X^{new})$.
\STATE Create transparent model $M^{new}$ based on $X^{*, new}$.
\end{algorithmic}
\end{algorithm}

\section{Empirical Study of SAFE}
\label{SAFE_application}

We performed benchmark on the selected data sets from OpenML100 \citep{bischl2017openml} collection of data sets for classification problems. We have selected binary classification data sets that do not contain missing values, in total 30 data sets.
Each data set in the OpenML100 is linked to a task defined in the OpenML database \citep{OpenML2013}. Each task provides 10 train/test splits and defined variable to be predicted. Some of the provided splits lead to the subsets of the original data set that contain variables with only one value. We have excluded such data sets due to the technical reasons.

For each train/test split in task we have trained 4 models: vanilla logistic regression, support vector machines on default hyperparameters (svm default), gradient boosting machines on default hyperparameters (gbm default), and tuned gradient boosting machines (gbm tuned). Hyperparameter tuning for gbm was performed on 20 randomly selected hyperparameter settings.  Models' hyperparameters and ranges for gbm tuning are in Table~\ref{tab:hyperparameters}. Values of tuned gbm hyperparameters differ between data sets.

\begin{table}[h]
\caption{Values of default gbm, values of default svm hyperparameters, and ranges of gbm hyperparameters used for tuning.}
\label{tab:hyperparameters}
\vskip 0.1in
\centering
\small
\begin{tabular}{ll}
\toprule
\textbf{hyperparemeter of default svm} & \textbf{value}  \\ 
\midrule
kernel       & Gaussian               \\
'C' - cost of constraints violation         & 1  \\
\bottomrule
\\
\toprule
\textbf{hyperparemeters of default gbm} & \textbf{values}  \\ 
\midrule
interaction depth       & 1               \\
number of trees         & 100  \\
shrinkage               & 0.1 \\
bag fraction            &  0.5 \\
\bottomrule
\\
\toprule
\textbf{hyperparemeters for gbm tunning} & \textbf{ranges}  \\ 
\midrule
interaction depth       & 1               \\
number of trees         & {[}50, 1000{]}  \\
shrinkage               & {[}0.01, 0.6{]} \\
bag fraction            & {[}0.2, 0.7{]} \\
\bottomrule
\end{tabular}
\vskip -0.1in
\end{table}

We used default svm, default gbm, and tuned gbm as surrogate models and applied SAFE method to extract new features. For changepoint penalty we used a Modified Bayes Information Criterion \citep{Zhang2006}. On the new features, we have trained logistic regression and obtained 3 new models: SAFE gbm default, SAFE gbm tuned, SAFE svm default.

\begin{figure}[!ht]
\vskip 0.2in
    \centering
    \includegraphics[width=\textwidth]{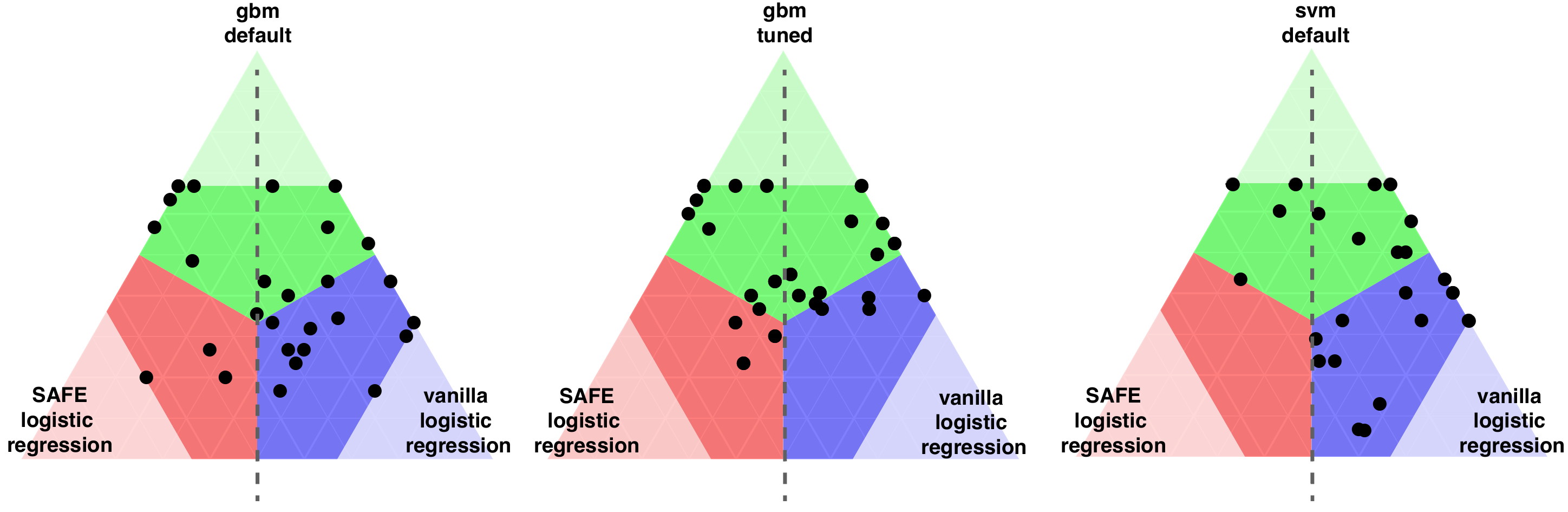}
    \caption{Ternary plots of AUC measures. One dot corresponds to models' performances on one data set. The position in the triangle is composed of AUC values of vanilla logistic regression, surrogate model, and refined logistic regression. Dots in the green area are data sets for which, on average, surrogate model was the best. Dots in the red area indicates data sets for which models refined with the SAFE method were the best, the blue area contains dots for which vanilla logistic regression was the best. On the left side of the vertical dashed line are data sets for which SAFE-based logistic regression models where on average better than vanilla logistic regression. The area marked by more saturated color shows a range of dots' possible appearance area. Dots cannot reach corner areas because their positions are calculated based on positions in AUC ranking among three models (baseline, surrogate, and refined). It means that for a split in a~data set, the best model gains 2 points, second gains 1 point, third gains 0 points. After averaging over 10 splits we obtain trinomial vector with averaged scores for three models. If the model would win on all splits, it would gain $\frac{2}{3}$ of the total sum of points, thereby not all parts of the triangle are reachable. Models that gain $\frac{2}{3}$ of the total sum of points lie on the one-colored edges of the hexagon.}
    \label{fig:triangle}
\end{figure}

We evaluated models with the AUC metric. A vanilla logistic regression without any feature extraction is considered as a baseline.
Complex models such as gbm and svm are surrogates required to perform SAFE algorithm. Refined models are logistic regressions trained on features extracted from SAFE method for different surrogate models. 
A list of used data sets, related OpenML tasks, and models' performances are in Table~\ref{tab:performance}. 
A way to visualize performances of triplets (vanilla logistic regression, surrogate model, refined SAFE model) is plotting them in barycentric coordinates, see Figure~\ref{fig:triangle}. 
In these plots we can distinguish two areas related to different kinds of results. 
\begin{itemize}
    \item The left side of the plot, separated by the vertical dashed line includes data sets where refined logistic regressions, on average, performed better than vanilla logistic regression. This corresponds to situations where extraction of information from complex models led to improving the performance of logistic regression models. The red area indicates data sets where SAFE-based logistic regression models performed better than complex surrogate models. The appearance of the data sets in the red area show that there are situations where appropriate feature engineering leads to simple model that achieves better performance than complex model. It may be surprising that the refined model is better than the surrogate one, however there are some reasons for that. Elastic models are better to capture non-linear relations but at the price of larger variance for parameter estimation. In some cases the refined models will work on better features and will have less parameters to train, thus it can outperform the surrogate model. This insight  questions a common myth that complex machine learning models out-perform linear ones.
    
    \item The right side of the plot, separated by the vertical dashed line includes data sets for which the vanilla logistic regression method achieved on average better performance than SAFE. Yet, blue area indicates the data sets where complex surrogate model was worse than vanilla logistic regression, therefore SAFE was unable to extract the variables from the complex model that will overtake the vanilla logistic regression performance.  Additionally, the appearance of data sets in the blue area show that despite the tuning, not every gbm model was able to achieve better results than the logistic regression.
\end{itemize}

\begin{table*}[ht]
    \caption{Mean AUC of models followed by standard deviation, calculated from 10 train/test splits defined for each data set in the OpenML database. 
    The highest values of AUC for each data set are bolded.}
    \label{tab:performance}
    \vskip 0.1in
    \centering
    \small
\begin{tabular}{c|ccccccc}
\toprule
\specialcell{\textbf{dataset}\\ \textbf{(OML task)}} & \specialcell{\textbf{vanilla logistic} \\ \textbf{regression}} & \specialcell{\textbf{gbm}\\ \textbf{default}} & \specialcell{\textbf{SAFE}\\ \textbf{gbm default}} & \specialcell{\textbf{gbm}\\ \textbf{tuned} }& \specialcell{\textbf{SAFE}\\ \textbf{gbm tuned}} & \textbf{svm} & \specialcell{\textbf{SAFE } \\ \textbf{svm}}\\
\midrule
credit-g (31) & \textbf{0.79+-0.04} & 0.78+-0.04 & 0.77+-0.04 & 0.78+-0.05 & 0.77+-0.03 & \textbf{0.79+-0.04} & 0.73+-0.05\\
diabetes (37) & 0.83+-0.06 & 0.83+-0.04 & \textbf{0.84+-0.04} & \textbf{0.84+-0.04} & 0.83+-0.04 & 0.83+-0.05 & 0.83+-0.04\\
spambase (43) & 0.97+-0.01 & \textbf{0.98+-0.01} & \textbf{0.98+-0.01} & \textbf{0.98+-0.01} & \textbf{0.98+-0.01} & \textbf{0.98+-0.01} & \textbf{0.98+-0.01}\\
tic-tac-toe (49) & \textbf{1.00+-0} & 0.81+-0.03 & 0.82+-0.04 & \textbf{1.00+-0} & 0.74+-0.05 & \textbf{1.00+-0} & 0.75+-0.05\\
electricity (219) & 0.75+-0.08 & 0.86+-0.01 & 0.86+-0.01 & \textbf{0.92+-0} & 0.86+-0.01 & 0.88+-0 & 0.84+-0.01\\
scene (3485) & 0.96+-0.02 & \textbf{0.98+-0.02} & 0.87+-0.03 & \textbf{0.98+-0.02} & 0.77+-0.02 & 0.94+-0.02 & 0.71+-0.03\\
\specialcell{monks-problems-1 \\ (3492)} & 0.70+-0.07 & 0.69+-0.06 & 0.70+-0.06 & 0.72+-0.06 & 0.72+-0.08 & \textbf{1+-0} & 0.71+-0.08\\
\specialcell{monks-problems-2 \\ (3493)} & 0.54+-0.10 & 0.54+-0.10 & 0.55+-0.11 & 0.53+-0.09 & 0.52+-0.07 & \textbf{0.65+-0.06} & 0.56+-0.10\\
\specialcell{monks-problems-3 \\ (3494)} & \textbf{0.99+-0.02} & 0.98+-0.03 & \textbf{0.99+-0.02} & \textbf{0.99+-0.02} & \textbf{0.99+-0.02} & 0.98+-0.03 & \textbf{0.99+-0.02} \\
\specialcell{gina\_agnostic (3891)} & 0.79+-0.02 & 0.92+-0.02 & 0.78+-0.03 & 0.94+-0.02 & 0.80+-0.03 & \textbf{0.96+-0.01} & 0.80+-0.03\\
mozilla4 (3899) & 0.89+-0.01 & 0.96+-0.01 & 0.90+-0.02 & \textbf{0.97+-0.01} & 0.89+-0.02 & 0.93+-0.01 & 0.91+-0.01\\
pc4 (3902) & 0.92+-0.03 & 0.93+-0.02 & 0.89+-0.03 & \textbf{0.94+-0.02} & 0.89+-0.03 & 0.90+-0.02 & 0.84+-0.05\\
pc3 (3903) & \textbf{0.82+-0.06} & \textbf{0.82+-0.03} & 0.78+-0.06 & \textbf{0.82+-0.04} & 0.79+-0.07 & 0.72+-0.08 & 0.79+-0.06\\
kc2 (3913) & 0.82+-0.12 & \textbf{0.85+-0.09} & 0.82+-0.09 & 0.84+-0.11 & 0.83+-0.11 & 0.78+-0.1 & 0.81+-0.12\\
kc1 (3917) & \textbf{0.80+-0.03} & \textbf{0.80+-0.04} & 0.79+-0.04 & \textbf{0.80+-0.04} & 0.79+-0.04 & 0.74+-0.06 & 0.79+-0.03\\
pc1 (3918) & 0.81+-0.07 & 0.82+-0.06 & 0.80+-0.07 & \textbf{0.83+-0.06} & 0.81+-0.09 & 0.78+-0.05 & 0.80+-0.08\\
\specialcell{MagicTelescope (3954)} & \textbf{1.00+-0} & \textbf{1.00+-0} & 0.99+-0 & \textbf{1.00+-0} & \textbf{1.00+-0} & \textbf{1.00+-0} & \textbf{1.00+-0}\\
wdbc (9946) & 0.95+-0.03 & \textbf{0.99+-0.01} & 0.97+-0.03 & \textbf{0.99+-0.01} & \textbf{0.99+-0.01} & \textbf{0.99+-0.01} & 0.96+-0.03\\
phoneme (9952) & 0.81+-0.02 & 0.87+-0.01 & 0.87+-0.02 & 0.90+-0.01 & 0.88+-0.01 & \textbf{0.91+-0.01} & 0.86+-0.01\\
qsar-biodeg (9957) & 0.92+-0.03 & 0.91+-0.03 & 0.91+-0.03 & 0.92+-0.03 & 0.91+-0.03 & \textbf{0.93+-0.03} & 0.90+-0.04\\
hill-valley (9970) & 0.59+-0.04 & 0.53+-0.04 & 0.55+-0.04 & \textbf{0.60+-0.06} & 0.58+-0.06 & 0.54+-0.07 & 0.53+-0.03\\
ilpd (9971) & \textbf{0.75+-0.07} & 0.73+-0.06 & 0.73+-0.05 & 0.73+-0.05 & 0.73+-0.07 & 0.66+-0.08 & 0.73+-0.08\\
madelon (9976) & 0.59+-0.04 & \textbf{0.69+-0.03} & 0.63+-0.03 & 0.68+-0.03 & 0.63+-0.04 & 0.62+-0.04 & 0.53+-0.01\\
\specialcell{ozone-level-8hr (9978)} & \textbf{0.90+-0.04} & 0.89+-0.04 & 0.89+-0.04 & \textbf{0.90+-0.03} & 0.88+-0.04 & \textbf{0.90+-0.04} & 0.83+-0.04\\
\specialcell{climate-model-\\simulation-crashes \\ (9980)} & \textbf{0.85+-0.1} & 0.82+-0.15 & 0.77+-0.1 & 0.81+-0.14 & 0.81+-0.11 & \textbf{0.85+-0.07} & 0.77+-0.08\\
\specialcell{eeg-eye-state (9983)} & 0.68+-0.01 & 0.78+-0.01 & 0.77+-0.01 & 0.85+-0.01 & 0.79+-0.01 & \textbf{0.88+-0.03} & 0.77+-0.01\\
\specialcell{banknote-authentication \\ (10093)} & \textbf{1.00+-0} & 0.99+-0.01 & 0.99+-0.01 & \textbf{1.00+-0} & \textbf{1.00+-0} & \textbf{1.00+-0} & 0.99+-0.01\\
\specialcell{blood-transfusion-\\service-center (10101)} & \textbf{0.75+-0.05} & \textbf{0.75+-0.05} & 0.74+-0.05 & \textbf{0.75+-0.05} & 0.74+-0.05 & 0.69+-0.05 & 0.71+-0.04\\
\specialcell{bank-marketing (14965)} & 0.91+-0.01 & 0.90+-0.01 & 0.89+-0.01 & \textbf{0.92+-0.01} & 0.88+-0.01 & 0.90+-0.01 & 0.89+-0.01\\
\specialcell{PhishingWebsites \\ (34537)} & \textbf{0.99+-0} & 0.98+-0 & 0.98+-0 & \textbf{0.99+-0} & 0.98+-0 & \textbf{0.99+-0} & 0.98+-0\\
\bottomrule
\end{tabular}
    \vskip -0.15in
\end{table*}

We supplemented performance of benchmarked models by their interpretability measures.
\citet{bertsimas2019price} showed that the trade-off between interpretability and performance is realized by decreasing performance with growing interpretability. What is more, they introduced the idea of measuring interpretability of linear models and trees. According to their approach, we assess the interpretability of the models by the inverse of number model's parameters. 
For linear regression models the number of parameters is the number of model's coefficients, including intercept.
 For svm models the number of parameters is the number of support vectors.  
 For gbm models the number of parameters is number of trees multiplied by 4. Each tree has maximum depth equals 1, therefore one parameter is selected variable in the node, second one is threshold for this variable, the last two are weights of two child nodes.

\begin{figure*}[!ht]
\vskip 0.2in
    \centering
    \includegraphics[width=\textwidth]{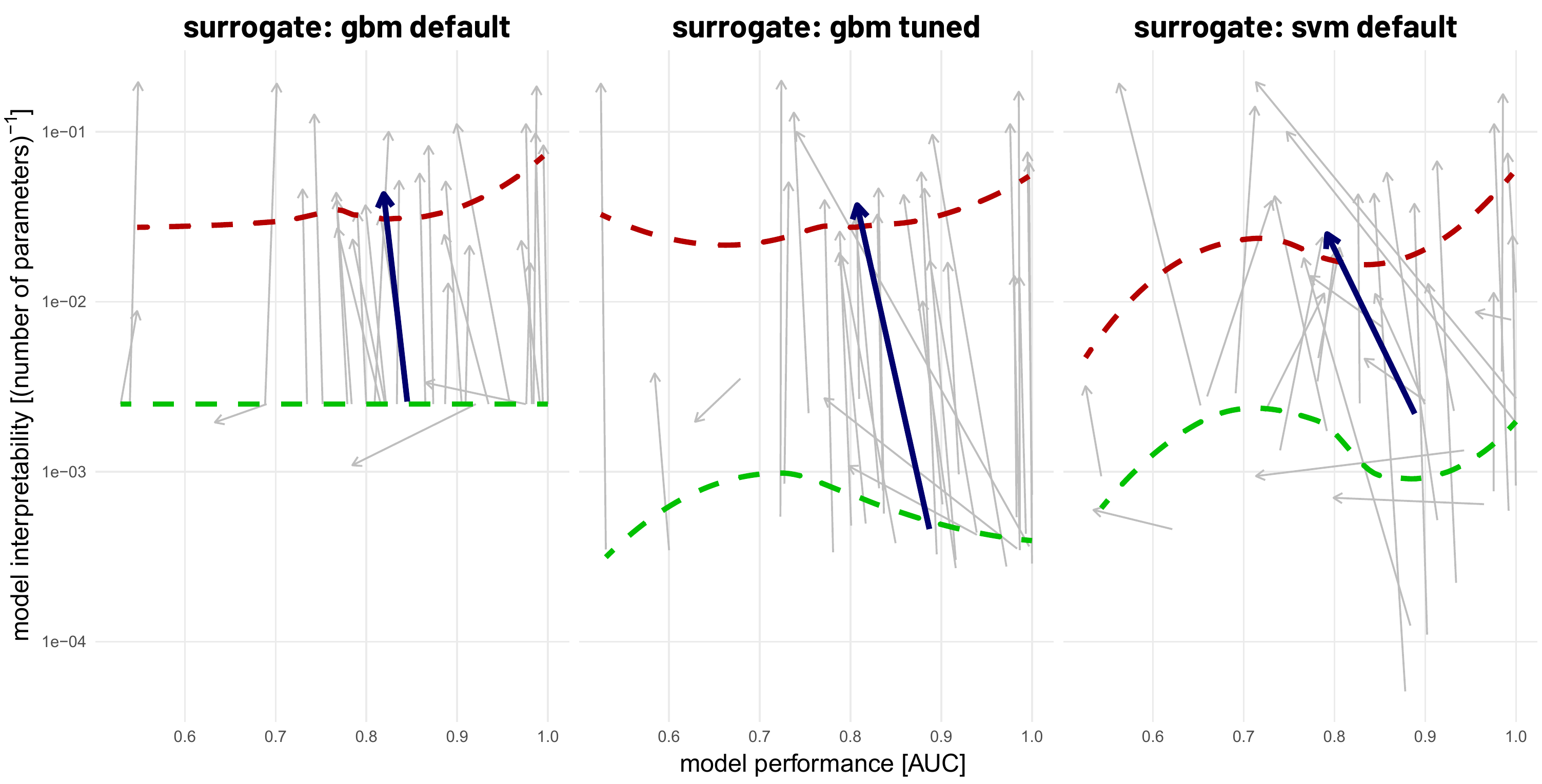}
    \caption{The interpretability-performance trade-off. The beginnings of grey arrows mark complex surrogate models' performances and their interpretability levels, the arrowheads mark SAFE-based refined models' performances and their interpretability levels. Therefore, grey arrows illustrate interpretability-performance shifts for data sets when using SAFE. The dark blue arrows shows medians. The Green dashed lines are interpretability trends for surrogate models, the red dashed lines are interpretability trends for SAFE-based refined models. Vertical offsets between these lines shows that SAFE lifted  the interpretability-performance trade-off.}
    \label{fig:tradeoff}
\end{figure*}

 The trade-off between interpretability and performance is shown in Figure~\ref{fig:tradeoff}. Median dark blue arrows shows the overall shift of the interpretability and performance after applying the SAFE method. We have used the Wilcoxon rank sum tests to check whether there are significant differences in AUC and interpretability between results of complex surrogate models and SAFE-based models. The p-values are in Table~\ref{tab:test_auc}. Tests show that in general there is no significant decrease of AUC after using SAFE and there is a significant increase of interpretability.

\begin{table}[]
    \label{tab:test_auc}
    \caption{The p-values of the Wilcoxon rank sum tests for equality of AUC values and for equality of interpretability levels. All p-values for equality of AUC tests are above the significance level 0.05, therefore we cannot reject the hypothesis about equality of AUC. The all p-values for equality of interpretability levels are below the significance level 0.05 which means that we can reject the null hypothesis in favor of the alternative hypothesis that interpretability levels are~different.}
    \vskip 0.1in
    \centering
    \small
    \begin{tabular}{lccc}
    \toprule
    \multicolumn{1}{c}{}    & \multicolumn{1}{c}{\specialcell{gbm default \\vs\\ SAFE}} & \multicolumn{1}{c}{\specialcell{gbm tuned \\vs\\ SAFE}} & \multicolumn{1}{l}{\specialcell{svm \\vs\\ SAFE}} \\ \midrule
    \specialcell{$H_0$: AUC values are equal }  & 0.458  & 0.119  & 0.109  \\   
    \vspace{0.5em} \\
     \specialcell{$H_0$: interpretability levels  are equal } & \specialcell{7.47e-13 \\ ***}  & \specialcell{7.61e-20 \\ ***}  &  \specialcell{1.43e-13\\ ***}                    \\ 
    \bottomrule
    \end{tabular}
\end{table}

Refined SAFE-based models are simple, with a small number of parameters, therefore one could conclude that refined models generalize data better than complex ones. However, it is worth noting that the refined models generalize relationships that were captured by surrogate models.
Thus, without a complex model as a surrogate, it would not have been possible.
With SAFE method, transferring knowledge about relationships to a simple model is automatic and do not require detailed investigation of the complex model.
Even if black-box model gains better results, it is still worth considering to apply transparent glass-box model. As we have seen in previous examples, performance of surrogate and refined model were, in general, close to each other. The advantage of a simpler model is that we gain transparency, interpretability and auditability.

\subsection{Example of Interpretability}

\begin{figure*}[!ht]
    \centering
    \includegraphics[width=0.9\textwidth]{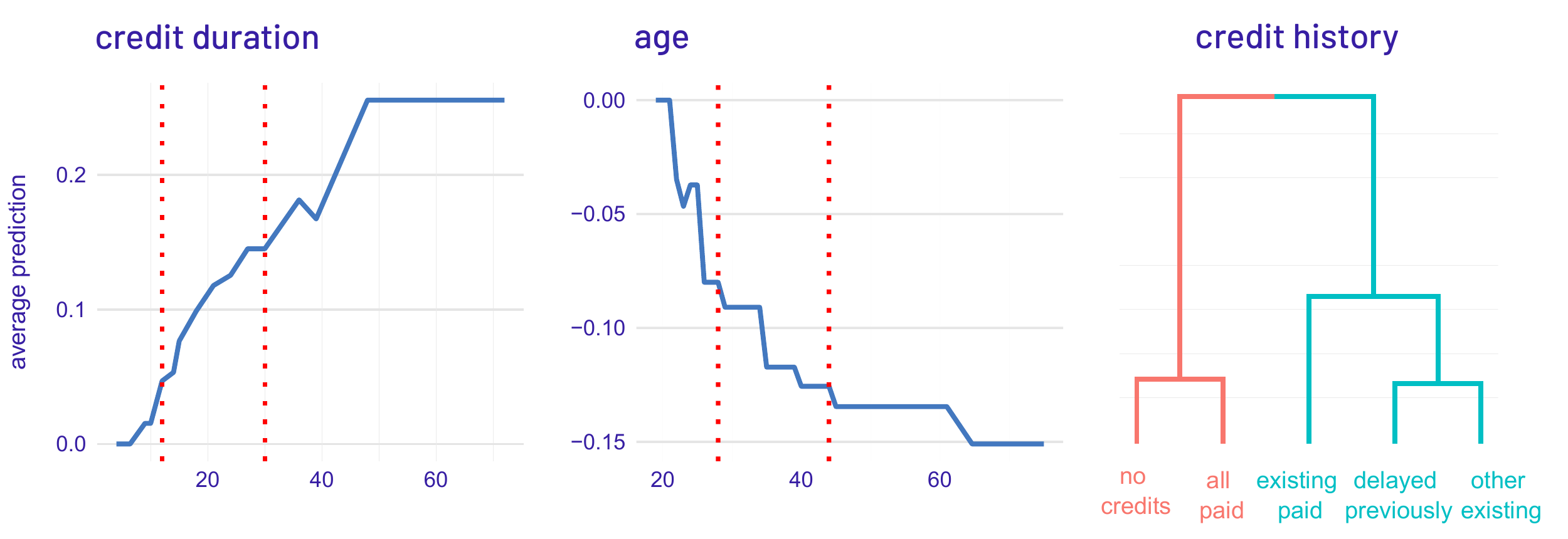}
    \caption{Example transformations of two continuous variables, credit duration and age, and a categorical variable credit history.}
    \label{fig:variableTransforamtion}
\end{figure*}

To explain interpretability of refined SAFE-based model we show example transformations for data set credit-g \citep{Dua:2017}, more detailed, for split 4 in task 31 from OpenML database. The data set classifies people described by a set of attributes as good or bad credit risks. Here AUC of vanilla logistic regression equals 0.78, while AUC of refined logistic regression equals 0.79. What is more, number of parameters for vanilla logistic regression is 49, while for refined logistic regression is 25. Therefore, we achieve better AUC, at the same time decreasing the number of parameters.
In Figure~\ref{fig:variableTransforamtion} are example SAFE transformations calculated for tuned gbm model. Coefficients od refined logistic regression for variables included in the Figure~\ref{fig:variableTransforamtion} are in Table~\ref{tab:variableTransformation}.
Refined logistic regression is fitted on new, binary features, therefore we can interpret coefficients in terms of probability. For example, credit duration value in $(12, 30]$ increases the logit of probability by 0.58.
The surrogate tuned gbm model achieved AUC equals 0.79, which is the same value as for refined logistic regression, yet with the logistic model we have gained higher interpretability.

\begin{table}[!ht]
\caption{Coefficients of logistic regression fitted to features transformed with SAFE method based on tuned gbm model on g-credit data set. We show only features included in Figure~\ref{fig:variableTransforamtion}.}
\label{tab:variableTransformation}
\vskip 0.1in
\centering
\small
\begin{tabular}{lc}
\toprule
\multicolumn{1}{c}{\textbf{feature}} & \multicolumn{1}{l}{\textbf{coefficient}} \\ 
\midrule
intercept  & -2.81    \\
credit duration (12, 30{]}   & 0.58    \\
credit duration (30, Inf)   & 1.11   \\
age (28, 44{]}   & -0.30  \\
age (44, Inf)   & -0.54     \\ 
credit history  \{ no credits, all paid \}  & 0.96 \\
\bottomrule
\end{tabular}
\vskip -0.1in
\end{table}

\newpage
\section{Discussion}
\label{discussion}

In this article, we presented a novel algorithm called SAFE that uses a machine learning surrogate model to automate feature transformations. New features are then used to train refined glass-box model, for example logistic regression. 
We benchmarked SAFE on 30 data sets from OpenML repository for classification problems. The results confirmed that SAFE algorithm produces features that can be further used to fit accurate and transparent model. We also justified the advantage of refined models over surrogate black-boxes, they are more interpretable and thus trustworthy.

The SAFE method allows us to fulfill four requirements of trustworthy predictive model, stated in Section~\ref{motivation}. One can choose a final refined model, accordingly to the simplicity and transparency, therefore statement (3) about interpretability is accomplished. Simple models, such as, linear regression and logistic regression are extensively described from a~mathematical point of view. As a result, there are many methods to diagnose such models. Therefore, requirement of the (2) auditability is also fulfilled. In Section~\ref{SAFE_application} we showed that performances of refined models are close to performance of complex surrogate models. Therefore, SAFE method allows to gain (1) high model performance. 
In Section~\ref{SAFE_application} we also argued that SAFE algorithm allows automatic feature transformation for the purpose of fitting refined model. This approach allows you to omit examining a~complex model. Thus (4) automaticity is \mbox{also accomplished.}

The results of a benchmark show that there are data sets where appropriate feature engineering may lead to fitting linear model that achieve equal or higher performance than complex models. These results confirm the value of extracting features from complex models in order to improve simple ones.

\subsection{Future Work}
\label{sec:future}

SAFE algorithm is used for transforming individual features. One can consider a natural extension of this approach to identification and extraction of interactions.
In this work the surrogate model could have any structure as the SAFE is model agnostic. For specific classes of surrogate models there are methods of capturing interactions. Most common approaches are developed for tree assembles like for random forest \cite{randomForestExplainer} or xgboost \cite{xgboostExplainer}.
This can be used for extraction of new features which contain information about interactions between variables.



\subsection{Software}
The benchmark was performed with the R package rSAFE (\url{https://github.com/ModelOriented/rSAFE}).
SAFE method is also implemented as a python library SafeTransformer (\url{https://github.com/ModelOriented/SAFE}).

\section*{Acknowledgements}
We would like to acknowledge Aleksandra Gacek and Piotr Luboń for developing Python library SAFE and Anna Gierlak for developing R package rSAFE.

We would like to thank Anna Kozak for the valuable discussions and beautification of the visual part of the paper.

Alicja Gosiewska was financially supported by the grant of the Polish Centre for Research and Development POIR.01.01.0100-0328/17. Przemyslaw Biecek was financially supported by the grant NCN Opus grant 2017/27/B/ST6/01307.

\bibliography{safe}

\end{document}